\documentclass[runningheads]{llncs}
\usepackage{graphicx}
\graphicspath{{./figures/}}

\usepackage{tikz}
\usepackage{comment}
\usepackage{amsmath,amssymb} 

\usepackage{color}

\usepackage[width=122mm,left=12mm,paperwidth=146mm,height=193mm,top=12mm,paperheight=217mm]{geometry}

\usepackage{subfigure}
\usepackage[hidelinks]{hyperref}
\usepackage{tabularx}
\usepackage{makecell,multirow,diagbox}

\usepackage{bbding}

\usepackage[marginal]{footmisc}

\begin{document}
\pagestyle{headings}
\mainmatter

\title{Infrared and visible image fusion based on Multi-State Contextual Hidden Markov Model} 


\titlerunning{MCHMM for Infrared and Visible Image Fusion}
%

\author{Xiaoqing Luo\inst{1} \and
Yuting Jiang\inst{1} \and
Anqi Wang\inst{1} \and
Zhancheng Zhang\inst{2(}\Envelope\inst{)} \and
Xiao-Jun Wu\inst{1}}
\authorrunning{X. Luo et al.}
%
\institute{ Jiangnan University, Wuxi, China \and
 Suzhou University of Science and Technology, Suzhou, China\\
 \email{cimszhang@163.com} 
 }

\maketitle


\begin{abstract}
The traditional two-state hidden Markov model divides the high frequency coefficients only into two states (large and small states). Such scheme is prone to produce an inaccurate sta- tistical model for the high frequency subband and reduces the quality of fusion result. In this paper, a fine-grained multi-state contextual hidden Markov model (MCHMM) is proposed for infrared and visible image fusion in the non-subsampled Shearlet domain, which takes full consideration of the strong correlations and level of details of NSST coefficients. To this end, an accurate soft context variable is designed correspondingly from the perspective of context correlation. Then, the statistical features provided by MCHMM are utilized for the fusion of high frequency subbands. To ensure the visual quality, a fusion strategy based on the difference in regional energy is proposed as well for low-frequency subbands. Experimental results demonstrate that the proposed method can achieve a superior performance compared with other fusion methods in both subjective and objective aspects.

\end{abstract}

\section{Introduction}
Infrared images can give the salient target information, while visible images contain fine visual information. The technique of infrared and visible image fusion which combine such unique and complementary characteristics has been widely used in security monitoring, foggy driving, resource detection and military applications. So far, a large quantity pf image fusion methods have been proposed based on different techniques\cite{2013Image,2015ALPSR,li2017multi,Luo2021IFSR,Luo2021Lat}, among them, multi-scale decomposition based methods (MSD) have become one of the most active field in image fusion, as the multi-scale decomposition mechanism is similar to human visual system, thus providing a high-level transform domain for the geometric abstraction and characterization of images \cite{1Ma2018Infrared}. Particularly, the recently proposed non-subsampled Shearlet transform (NSST) overcomes the disadvantages of previous MSD methods to realize the shift-invariance, fast decomposition and direction sensitivity, which improves the preservation of geometrical structure \cite{luo2016novel}. Therefore, NSST is chosen as the basis of our research.

Even though MSD-based image fusion methods have achieved state-of-the-art performance, still, there exist several challenges when determining the feature extracted from coefficients, which is a critical factor contributes to the effectiveness of subsequently manual-designed fusion rule. Simple features, e.g. energy, deviation and gradients, calculated directly from the MSD coefficients under the assumption of coefficient independence can only provide unilateral image presentation and degrade the visual quality of fused results. To obtain a more reliable fused image, it is necessary to analyze the hierarchy of image information for better adaption to the complex relationship between multi-modal images.
	
In the past few decades, statistical model has been proved with the benefits to exploit effective data presentation through modeling the probability of occurrence in images \cite{3Nuha2019A}.  Especially, combining the statistical model with MSD coefficients to capture their correlative features is able to increase the interpretability of intrinsic image patterns which cannot be simply defined by edges or lines. Due to their successful applications and great improvement of performance in computer vision related tasks, such as image retrieval and classification \cite{4Li2019Marginal}, statistical models have also been introduced to MSD-based image fusion methods to facilitate the design of activity level measurement and fusion rule \cite{5Zhang2020Statistical}, while how to formulate an reasonable joint distribution model is still a great obstacle as the multiple dependencies among coefficients, e.g. inter-scale, inner-scale, inter-direction, etc. Crouse et al. \cite{6Crouse1997} proposed the Contextual hidden Markov model (CHMM) to simultaneously capture the complex correlational relationships via a variable gleaning the context information. Nevertheless, the traditional practice in CHMM tends to use a two-state Gaussian mixture model (GMM) to model the distribution of MSD coefficients, which means that each coefficient is simply divided into the position with rich texture information and the smooth area \cite{7Xia2001Image,8Long2009Statistical,9luo2017image}. Such rough (two-state) model combined with a hard context variable fails to capture the diverse levels of detail amount contained in the coefficients. Therefore, some fine texture features may also be incorrectly classified into the small state, which leads to an inaccurate statistical model and further affects the quality of fused image. Song et al. \cite{10Song2002Document} once proposed a three-state hidden Markov tree (HMT) model for document segmentation, which assigns the wavelet coefficients into three states: background, text, and image, improving the segmentation accuracy compared with the two-state HMT based method. However, this method is only suitable for limited image application, and it is worth exploring the construction of a multi-state statistical model to cover the diverse situations of image coefficients.
Under the above consideration, a multi-state CHMM model (MCHMM) is constructed for infrared and visible image fusion in this paper. Firstly, the source images are decomposed by NSST into low frequency and high frequency subbands, and a multi-state GMM is used in MCHMM to estimate the distribution of NSST coefficients. The multiple states represent the different levels of detail, which can help get a more fine-grained model. Correspondingly, a soft context variable is proposed to measure the degree of detail of the coefficient through the exploration of multiple correlations between the coefficients, which further improves the interpretability of the statistical model. Then, a novel activity-level measure based on the combination of multi-state statistical characteristics of image and the soft context of MCHMM is designed for the fusion of high frequency subbands. The fusion of low frequency subbands adopts a weighted fusion rule based on the difference of regional energy. To sum up, the main contributions of this paper are fourfold:
\begin{enumerate}
	\item The MCHMM is established on the coefficients of NSST, for which multiple states are designed for the statistical model, which means that the details of the coefficients are defined in multiple levels to reflect the degree of the amounts of details, thus improving the precision and generalization performance of the proposed model.  
	\item A soft context scheme is designed. The soft context variable can make fine-grained evaluation for the details of coefficients from the perspective of intra-scale and inter-scale correlations. 
	\item A novel activity level measure of high frequency coefficient is constructed based on the multi-state statistical characteristics and the soft context variable, which can evaluate the saliency degree in association with the contribution of different detail level, facilitating a comprehensive and reliable fusion strategy to preserve the detail and structure better.
	\item The fusion of low frequency subbands depends on the difference of regional energy, which can help enhance the contrast of the fused image for the consistency with human visual perception.  
\end{enumerate}	

\section{Related Works}
In 1998, M. S. Crouse \cite{11Crouse1998Wavelet} pointed that the wavelet coefficients have certain correlation with each other, and developed a framework based on wavelet-domain hidden Markov models (HMM) according to the statistical correlation and non-Gaussian statistical characteristics of real signals, which is demonstrated with the generalized utility for signal processing task. Simoncelli \cite{12Simoncelli1999Modeling} also found that the wavelet coefficients of images have certain correlation with each other, and such potential relationship can be fully exploited by statistical models to help the understanding of the intrinsic structure of images. Therefore, many work has concentrated on the statistical characteristics of MSD coefficients to carry out the relevant research on image processing tasks. Jian et al.\cite{13Jian0Fusion}  used the Hidden Markov Tree model (HMT) to capture the intra-dependency of NSST correlations, and the trained coefficients by HMT are selected based on local energy of gradients, which can greatly enhance the details in fused image. Wang et al. \cite{14WANG2020105387} proposed a NSCT-HMT based on the Gaussian copula function to model the correlation between the current coefficient with its four neighbors. The application of copula entropy facilitating the localization of texture features in the source images, and provide good performance in image denoising. Although the HMT model is an effective statistical model, it can be extremely difficult and time-wasting to capture different kind of correlations, such as model the inter-scale and intra-scale correlations simultaneously. Thereby, Course et al. \cite{6Crouse1997} proposed the contextual hidden Markov model(CHMM), which predicts the coefficients based on the hidden states conditioned on a context variable. Through designing the context variable, the information of different correlation on the hidden state can be gathered and modelled by HMM in a simple way. Fan et al. \cite{15Fan2001Image} proposed a local contextual hidden Markov model (LCHMM) for image denoising by utilizing the local statistical characteristics of wavelet coefficients and the correlation within the scale, and the ideal is further developed by Long et al. \cite{8Long2009Statistical} to the Contourlet domain. Luo et al. \cite{9luo2017image} proposed a CHMM to model the inter-direction, inter-scale and intra-scale correlations of NSST coefficients, and the contextual statistical similarity is proposed to measure the contribution of each coefficient to the fused result, leading to a reliable fusion strategy for redundant and complementary information. Zhang et al. \cite{16Zhang2020Multimodal} combined the global, local and regional CHMM model to represent the NSST coefficients in a more comprehensive way, and the global-regional-local statistical feature is extracted to measure the activity level of coefficients. The fused results for multi-modal images demonstrated their superiority in detail preservation. Although the above NSST based image fusion methods have achieved remarkable results with the adoption of statistical model, the detail level of coefficient is limited to small and large, which apparently contradicts the hierarchy of information and effects the accuracy in determining the transferred features. To better reflect the possibility of the saliency degree for each coefficient, we focused on the design of a multi-grained joint distribution model based on NSST to deal with the uncertainty in fusion process. 

\section{The framework and procedure of the proposed method} \label{sec:propsed_method}

The fusion framework is shown in Fig. \ref{fig:framework}. The specific steps of the algorithm are as follows:

\begin{figure}[htb]
	\def\tempwidth{0.8\textwidth}
	\tiny
	\centering
	\subfigure{\includegraphics[width=\tempwidth]{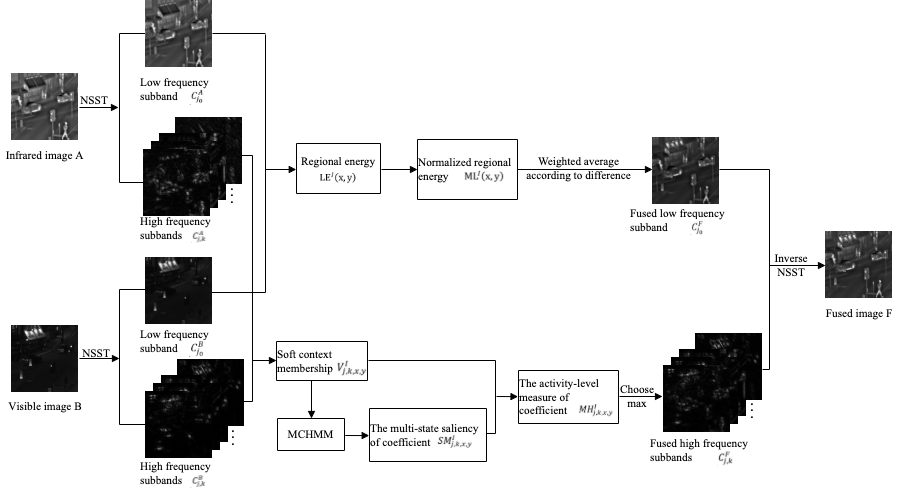}}
	\centering
	\caption{Fusion framework of the proposed algorithm.}
	\label{fig:framework}
\end{figure}

1) Assuming that the infrared image $A$ and the visible image $B$ to be fused are registered, NSST is performed on them to obtain high-frequency coefficients $C_{j,k}^A(x,y)$,$C_{j,k}^B(x,y)$ and low-frequency coefficients$C_{j_0}^A(x,y),C_{j_0}^B(x,y)$ respectively. Where $C_{j,k}^I(x,y)$ represents the high frequency coefficient located at $(x,y)$ position in the k-th direction of the j-th level, and $C_{j_0}^I(x,y)$ represents the low frequency coefficient located at the $(x,y)$ position in the coarsest scale $j_0$, $I$ represents the source image $A$ or $B$.

2) Extract the regional energy $LE^I(x,y)$ of the low-frequency subband coefficient, and obtain the low-frequency fusion subband $C_{j_0}^F$ based on the weighted fusion rule with the difference of the normalized regional energy $ML^I(x,y)$.

3) The high frequency subbands are modeled by NSST-MCHMM with the soft context scheme. The NSST-MCHMM is trained by the expectation maximization (EM) algorithm to get the multi-state contextual statistical parameters, which are used to calculate the multi-state saliency of coefficient ${SM}_{j,k,x,y}^I$. The fused high-frequency subband $C_{j,k}^F$ is obtained with the choose-max fusion rule based on the activity-level measure of coefficients ${MH}_{j,k,x,y}^I$, which consists of the multi-state saliency ${SM}_{j,k,x,y}^I$ and the soft context variable ${j,k,x,y}^I$.

4) The fused image $F$ is achieved by the inverse NSST transform.

\section{The Multi-State Contextual Hidden Markov Model for the NSST High- frequency Subband (NSST-MCHMM)}

In reference \cite{9luo2017image}, it has been verified that the distribution of NSST coefficients is non-Gaussian and the coefficients are correlated with each other. Traditionally, the distribution of coefficients is regarded as the two-state zero-mean Gaussian mixture distribution (GMM) in CHMM, where each point is simply defined as a large state or a small state. Moreover, the binary context variable is used in CHMM. In this case, the fine detail information of coefficients is easily ignored in the rough statistical model, which leads to the extraction of false features and a sub-optimal fusion result. It is necessary to enhance the granularity of statistical model to capture the subtle details. Therefore, a novel multi-state contextual hidden Markov model which model the multiple correlation of NSST coefficients is proposed to describe the saliency of coefficient more accurate and represent the image in a more comprehensive way.

\subsection{The NSST Transform}
\begin{figure}[htb]
	\def\tempwidth{0.8\textwidth}
	\tiny
	\centering
	\subfigure{\includegraphics[width=\tempwidth]{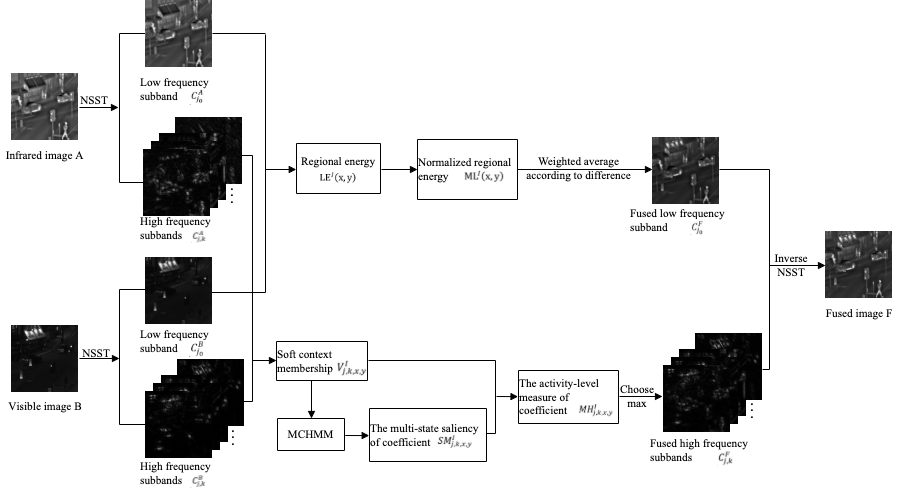}}
	\centering
	\caption{The decomposition of NSST.}
	\label{fig:nsst}
\end{figure}

Nowadays, NSST is widely used in image recognition, detection, denoising, fusion and other fields due to its low computational complexity and the advantages of multi-direction and translation invariance. The implementation of NSST \cite{17Liu2019Multi} can be divided into two steps of multi-scale and multi-directional decomposition, as shown in Fig. \ref{fig:nsst}.  First, the image is decomposed by the 2-level non-downsampling pyramid (NSP). There are two high frequency subbands and a low frequency subband. Since there is no down-sampling operation, the obtained subband images have the same size as the source image, which avoids the pseudo-Gibbs phenomenon at the singular point of image. Then, the high-frequency subbands are decomposed by the shearlet filter bank (SF) to obtain high-frequency direction subbands. This step ensures that the NSST has directional sensitivity. Therefore, the NSST can represent images well in terms of scale and orientation.

\subsection{Design scheme of soft context}
Commonly, the coefficient is defined as two states in the traditional CHMM model, in which “1” means the detail state (large state) and “0” means the smooth state (small state). Accordingly, the context variable is also defined as two-state. To adapt to the design of multi-state in the MCHMM model, a new soft context scheme is proposed. In this proposed scheme, the context variable is continuous with the range in [0\ ,1]. The value of context variable is regarded as an contextual information indicator under the exploration of inter-scale and intra-scale correlations of coefficients. The design of context is as follows:

\begin{small}
\begin{equation}
context=\omega_0\cdot\sum_{t=1}^{4}\left|{NA}_t\right|^2+\omega_1\cdot\sum_{t=1}^{4}\left|{NB}_t\right|^2+\omega_2\cdot\left|PX\right|^2+\omega_3\cdot\left(\left|{CX}_1\right|^2+\left|{CX}_2\right|^2\right),
\label{eq:1}
\end{equation}	
\end{small}
where NA and NB denote the direct and diagonal neighbors of the current coefficient respectively, PX represents the parent coefficient, and CX1 and CX2 represent two cousin coefficients. Under this design, the soft context variable $V$ is determined as follows and the illustration of $V$ is shown in Fig. \ref{fig:V}:

\begin{figure}[htb]
	\def\tempwidth{0.5\textwidth}
	\tiny
	\centering
	\subfigure{\includegraphics[width=\tempwidth]{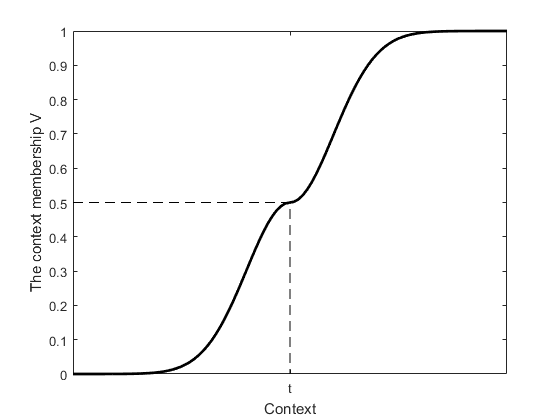}}
	\centering
	\caption{The graph of $V$.}
	\label{fig:V}
\end{figure}

\begin{equation}
    V=\left\{\begin{matrix}\frac{1}{2}\exp{\left(-\frac{1}{2}\left(\frac{context-t}{\sigma}\right)^2\right)},context<t\\\frac{1}{2},context=t\\1-\frac{1}{2}\exp{\left(-\frac{1}{2}\left(\frac{context-t}{\sigma}\right)^2\right)},context>t\\\end{matrix}\right.,
    \label{eq:2}
\end{equation}

\begin{equation}
    {t=4\cdot\omega}_0\cdot E+4\cdot\omega_1\cdot E+\omega_2\cdot E_P+\omega_3\cdot\left(E_{C_1}+E_{C_2}\right),
    \label{eq:3}
\end{equation}

where $\sigma$ represents the standard deviation of the context subband, $\omega_0,\omega_1,\omega_{2,}\omega_3$ are the weight factors of NA, NB, PX, CX1 and CX2, $E,\ E_P,\ E_{C_1},\ E_{C_2}$ represent the average energy of current subband, parent subband and two nearby cousin subbands respectively, which are calculated as 
\begin{equation}
    E_\ast=\frac{1}{N}\cdot\sum_{i=1}^{N}C^2\left(i\right),
    \label{eq:4}
\end{equation}
where $E_\ast$ stands for the four mean values $\left\{E,E_P,E_{C_1},E_{C_2}\right\}$, $N$ and $C^2$ are the total number of coefficients and the square of coefficients in the corresponding subband respectively. As can be seen from Fig. \ref{fig:V}, when context is equal to $t$, the probability that the coefficient belongs to the detail is 0.5. This probability increases as the value of $V$ increases. On the contrary, when the value of $V$ is smaller, the coefficient at the point is more likely to be smooth, and the probability of detail is lower.

\begin{figure}[htb]
	\def\tempwidth{0.8\textwidth}
	\tiny
	\centering
	\subfigure{\includegraphics[width=\tempwidth]{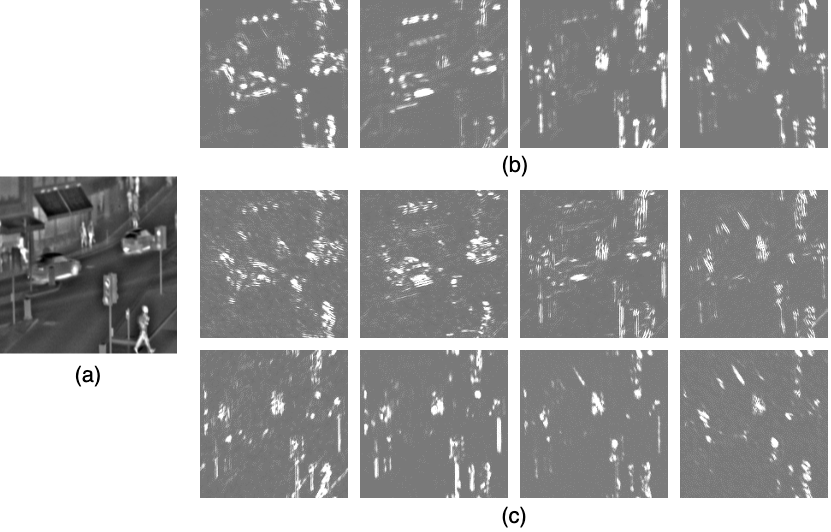}}
	\centering
	\caption{The visualization of V. (a) the visible image; (b) the context maps in four directions high-frequency subbands of the first scale; (c) the context maps in eight directions high-frequency subbands of the second scale.}
	\label{fig:VoV}
\end{figure}

The visualization of the soft context variables of the high-frequency subbands obtained by the two-level NSST decomposition for Fig. \ref{fig:VoV} (a) is shown in Fig. \ref{fig:VoV}. The value of the $V$ is projected into the gray space and the size of the gray value corresponds to the strength of the coefficient details. When the pixel is dark, this means that the value of $V$ is 0 and the corresponding coefficient is in the smooth region. When the pixel is white, this means that the value of $V$ is 1 and the corresponding coefficient has rich details. The source image Fig. \ref{fig:VoV}(a) is an infrared image that contains target information such as pedestrians, vehicles, and traffic lights. It is observed that the corresponding positions of targets show high brightness in Fig.\ref{fig:VoV} (b-c). This indicates that V can accurately reflects the detailed information of the source image. Therefore, the proposed soft context is effective in gleaning precise information based on the correlations of coefficients.

\subsection{The construction of NSST-MCHMM}
The NSST-MCHMM is based on multi-state GMM. Multi-state means that the Gaussian distributions with different variances are used to estimate the distribution of coefficients. The coefficients are divided into different levels of detail, which is conducive to fitting the distribution of coefficients and obtaining a fine-grained statistical model. The MCHMM model is established based on an n-state, zero-mean GMM, and each high frequency coefficient is associated with a soft context variable and a hidden state. The NSST-MCHMM can be established and collectively expressed by Eq. (\ref{eq:5}) and Eq. (\ref{eq:6}).

\begin{equation}
\begin{split}
   & f_{C_{j,k,x,y}|V_{j,k,x,y}}(C_{j,k,x,y}|V_{j,k,x,y}=v)=\\
   & \sum_{m=0}^{n-1}{P_{S_{j,k,x,y}|V_{j,k,x,y}}(S_{j,k,x,y}=m|V_{j,k,x,y}=v)}\cdot g\left(C_{j,k,x,y};0,{\sigma^2}_{j,k,x,y,m}\right)  ,
  \end{split}
    \label{eq:5}
\end{equation}

\begin{equation}
\begin{aligned}
   & P_{S_{j,k,x,y}|V_{j,k,x,y}}\left(S_{j,k,x,y}=m\middle| V_{j,k,x,y}=v\right)=\\
&   \frac{P_{S_{j,k,x,y}}(S_{j,k,x,y}=m) \cdot P_{V_{j,k,x,y|S_{j,k,x,y}}}(V_{j,k,x,y}=v|S_{j,k,x,y}=m)}{\sum_{m=0}^{n-1}{P_{S_{j,k,x,y}}(S_{j,k,x,y}=m)}\cdot P_{V_{j,k,x,y}|S_{j,k,x,y}}(V_{j,k,x,y}=v|S_{j,k,x,y}=m)}
 \end{aligned}
    \label{eq:6}
\end{equation}

The definition of parameter set is as follows:
\begin{small}
\begin{equation}
\begin{aligned}
&\mathrm{\Theta}= \\
&\left\{P_{S_{j,k,x,y}}\left(m\right),{\sigma^2}_{j,k,x,y,m},P_{V_{j,k,x,y}|S_{j,k,x,y}}\left(v\middle| m\right)|m=0,\cdots,n-1,v\in[0,1]\right\}
\end{aligned}
    \label{eq:7}
\end{equation}
\end{small}

Here, $n$ is a positive integer greater than or equal to 2, indicating the total number of states in the MCHMM; $C_{j,k,x,y}$ represents a random variable associated with the current coefficients $C_{j,k}(x,y)$ ;  $V_{j,k,x,y}$ and  $S_{j,k,x,y}$ denote the soft context variable and the hidden state variable of $C_{j,k,x,y}$ respectively; $P_{S_{j,k,x,y}|V_{j,k,x,y}}$ is the probability that the hidden state $S_{j,k,x,y}$ falls in the state $m$ when the value of context variable is $v$; $g\left(C_{j,k,x,y};0,{\sigma^2}_{j,k,x,y,m}\right)$ is the PDF (zero-mean Gaussian distribution with variance ${\sigma^2}_{j,k,x,y,m}$) for the $C_{j,k,x,y}$ conditioned on the state m.

\subsection{The EM training algorithm}
The NSST-MCHMM is trained by Expectation Maximization (EM) algorithm \cite{18Zhang2014Statistical}. The complete algorithm includes two stages. The first stage is the initialization stage, and the second stage is the iterative training part. The details are as follows.

1. Initialize the parameters 

(1) Set initial parameters as 

The definition of parameter set is as follows:
\begin{equation}
	P_{S_{j,k}}\left(m\right)=\frac{1}{n}; {\sigma^2}_{j,k,m}=\frac{\left[2{\delta^2}_{j,k}-{\sigma^2}_\eta-{\sigma^2}_\eta\ \right]}{n-1}\cdot m+{\sigma^2}_\eta
    \label{eq:8}
\end{equation}
where ${\sigma^2}_\eta$ is the known noise variance, and $\delta_{k,l}^2$ is the average energy of $C_{j,k,x,y}$. 

(2) Expectation (E) step. According to Bayes’ theorem, the following probabilities are calculated for each coefficient in global scope:
\begin{equation}
P(S_{j,k,x,y}=m|C_{j,k,x,y})={\frac{P_{S_{j,k}}\left(m\right)\ast g(C_{j,k,x,y};0,{\sigma^2}_{j,k,m})}{\sum_{m=0}^{n-1}{P_{S_{j,k}}(m)}\ast g(C_{j,k,x,y};0,{\sigma^2}_{j,k,m})}}.  
\label{eq:9}
\end{equation}

(3) Maximization (M) step. With the obtained probability in step E, the other parameters are calculated as follows:
\begin{equation}
P_{S_{j,k}}\left(m\right)=\frac{1}{M_{j,k}\cdot N_{j,k}}\sum_{x=1}^{M_{j,k}}\sum_{y=1}^{N_{j,k}}{P\left(S_{j,k,x,y}=m\ \middle| C_{j,k,x,y}\right)}
\label{eq:10}
\end{equation}

\begin{equation}
P_{S_{j,k}}\left(m\right)=\frac{1}{M_{j,k}\cdot N_{j,k}\cdot P_{S_{j,k}}(m)}\sum_{x=1}^{M_{j,k}}\sum_{y=1}^{N_{j,k}}{{C^2}_{j,k,x,y}\cdot P(S_{j,k,x,y}=m|C_{j,k,x,y})},
\label{eq:11}
\end{equation}
where $M_{j,k}$ and $N_{j,k}$ are the number of rows and columns of $C_{j,k,x,y}$, respectively.

(4) Increase the training times and go back to step (2) until the parameters converge or the number of iterations reaches the preset limit, then move on to the next step.

(5) Set the window of size $\left(2W_j+1\right)\times\left(2W_j+1\right)$ and use the following formulas to get the local initialization parameters from the global initialization results. Set a parameter $T$ as model training times and its initial value is defined as zero.
\begin{equation}
P_{S_{j,k,x,y}}\left(m\right)=\frac{1}{\left(2W_j+1\right)^2}\sum_{i=x-W_j}^{x+W_j}\sum_{l=y-W_j}^{y+W_j}{P\left(S_{j,k,i,l}=m\middle| C_{j,k,i,l}\right)}, 
\label{eq:12}
\end{equation}

\begin{equation}
\begin{split}
   & {\sigma^2}_{j,k,x,y,m}=\\
  & \frac{1}{{(2W_j+1)}^2\cdot P_{S_{j,k,x,y}}(m)}\sum_{i=x-W_j}^{x+W_j}\sum_{l=y-W_j}^{y+W_j}{{C^2}_{j,k,i,l}\cdot P(S_{j,k,i,l}=m|C_{j,k,i,l})}, 
\end{split}
\label{eq:13}
\end{equation}

\begin{small}
\begin{equation}
\begin{split}
    & P_{V_{j,k,x,y}|S_{j,k,x,y}}\left(v|m\right)=\\
& \frac{1}{{(2W_j+1)}^2\cdot P_{S_{j,k,x,y}}(m)}\sum_{i=x-W_j}^{x+W_j}\sum_{l=y-W_j}^{y+W_j} w\left(i,l\right) \cdot P(S_{j,k,i,l}=m|C_{j,k,i,l},V_{j,k,i,l}=v).
\end{split}
\label{eq:14}
\end{equation}
\end{small}

Here, $w\left(i,l\right)=exp\left(-\frac{{(v-c)}^2}{{2\sigma}^2+\varepsilon}\right)$, where $v$ represents the soft context variable of each coefficient in the local window, $c$ is the soft context variable of the current coefficient, $\sigma^2$ is the variance of soft context variable in the local window, and $\varepsilon$ represents a constant to avoid the denominator being zero. In our scheme, w indicates the context weight of the neighborhood coefficients in the local window.

2. Iterative EM Training.

(1) Expectation (E) step. Calculate the probability for each coefficient as follow:

\begin{equation}
\begin{aligned}
& P_{S_{j,k,x,y}|V_{j,k,x,y},C_{j,k,x,y}}\left(m\middle| C_{j,k,x,y},V_{j,k,x,y}=v\right)= \\
& {\frac{P_{S_{j,k,x,y}}\left(m\right)P_{V_{j,k,x,y}|S_{j,k,x,y}}\left(v\middle| m\right)\ast(g(C;0,{\sigma^2}_{j,k,x,y,m})}{\sum_{m=0}^{n-1}{P_{S_{j,k,x,y}}\left(m\right)}P_{V_{j,k,x,y}|S_{j,k,x,y}}(v|m)\ast(g(C;0,{\sigma^2}_{j,k,x,y,m})}}.
\end{aligned}
\label{eq:15}
\end{equation}

(2) Maximization (M) step. The parameters of the model are updated by the following formulas:
\begin{equation}
\begin{split}
P_{S_{j,k,x,y}}(m)=\frac{1}{{(2W_j+1)}^2}\sum_{i=x-W_j}^{x+W_j}\sum_{l=y-W_j}^{y+W_j} P_{S_{j,k,i,l}|V_{j,k,i,l}}(m|C_{j,k,i,l},V_{j,k,i,l})
\end{split}
\label{eq:16}
\end{equation}

\begin{small}
\begin{equation}
\begin{split}
{\sigma^2}_{j,k,x,y,m}=\frac{1}{{(2W_j+1)}^2\cdot P_{S_{j,k,x,y}}(m)}\sum_{i=x-W_j}^{x+W_j}\sum_{l=y-W_j}^{y+W_j}{{C^2}_{j,k,i,l}P(S_{j,k,i,l}=m|C_{j,k,i,l})}, 
\end{split}
\label{eq:17}
\end{equation}
\end{small}

\begin{small}
\begin{equation}
\begin{aligned}
    & P_{V_{j,k,x,y}|S_{j,k,x,y}}\left(v|m\right)=\\
& \frac{1}{{(2W_j+1)}^2\cdot P_{S_{j,k,x,y}}(m)}\sum_{i=x-W_j}^{x+W_j}\sum_{l=y-W_j}^{y+W_j}{w(i,l)}\cdot P_{S_{j,k,i,l}|V_{j,k,i,l}}(m|C_{j,k,i,l},V_{j,k,i,l}=v)
\end{aligned}
\label{eq:18}
\end{equation}
\end{small}

The calculations are performed on the window centered on $C_{j,k,x,y}$. Set training times $T=T+1$, and iterate the EM training step until the parameters converge or reach the preset maximum number of iterations. 

\section{The proposed fusion method}

\subsection{The Fusion Rule of Low Frequency Subband}
The low-frequency subbands contain the approximate information of image. The traditional choose-max or average fusion rule cannot effectively combine the useful low-frequency information. The visible image can reflect the scene with fine visual details, but shows low contrast between target and background. On the contrary, the infrared image which presents the thermal radiation of scene can provide salient target information, but it is weak in detail expression. Based on the different characteristics of infrared and visible image, a novel fusion rule based on the difference of regional energy is proposed to preserve the low-frequency details while maintaining the contrast of image. The fusion rule is defined as 

\begin{equation}
\begin{split}
C_{j_0}^F\left(x,y\right)=w\left(x,y\right)\ast C_{j_0}^A\left(x,y\right)+(1-w\left(x,y\right))\ast C_{j_0}^B\left(x,y\right),  
\end{split}
\label{eq:19}
\end{equation}

\begin{equation}
\begin{split}
w\left(x,y\right)=0.5+({\rm ML}^A\left(x,y\right)-{\rm ML}^B\left(x,y\right))/2, 
\end{split}
\label{eq:20}
\end{equation}
where $w\left(x,y\right)$ is the weight of $C_{j_0}^A\left(x,y\right)$, and the weight of $C_{j_0}^B\left(x,y\right)$ is defined as  $1-w\left(x,y\right)$. The weight is adaptively determined by the regional energy of the low-frequency coefficients as follows:
\begin{equation}
\begin{split}
{\rm ML}^I\left(x,y\right)=\left({\rm LE}^I\left(x,y\right)-min\left({\rm LE}^I\right)\right)/\left(max\left({\rm LE}^I\right)-min\left({\rm LE}^I\right)\right),
\end{split}
\label{eq:21}
\end{equation}

\begin{equation}
\begin{split}
{\rm LE}^I\left(x,y\right)=\frac{1}{W^2}\sum_{i=-(W-1)/2}^{(W-1)/2}\sum_{l=-(W-1/2)}^{(W-1)/2}{C_{j_0}^I(x+i,y+l)}^2,  
\end{split}
\label{eq:22}
\end{equation}
where ${\rm LE}^I\left(x,y\right)$ represents the regional energy of the low-frequency coefficient, $ I\ \epsilon \left\{A,\ B\right\}$; ${\rm ML}^I\left(x,y\right)$ is the normalized regional energy, and $W$ is the size of local window. If the difference between corresponding low frequency coefficients is great, then the low-frequency coefficient with higher regional energy has more contribution to the fused image, and if ${\rm ML}^A\left(x,y\right)$ is equal to ${\rm ML}^B\left(x,y\right)$, then they get the same fusion weight.

\subsection{The Fusion Rule of High Frequency Subbands}
The high-frequency subband contains the detailed information of the source image, so the fusion of high-frequency subbands is very important. To obtain an effective fusion result, a more accurate activity-level measure should be designed. In this paper, statistical correlation features of high frequency subbands are extracted from MCHMM model, which are utilized to define multi-state saliency of coefficient. The activity-level measure of coefficient is defined by its multi-state saliency and soft context variable. The specific fusion processing is described as follows.
First, combine the variance with the statistical feature to measure the detailed information ${VP}_{j,k,x,y,m}^I$, which is expressed as follow:
\begin{small}
\begin{equation}
\begin{split}
{VP}_{j,k,x,y,m}^I={Var}_{j,k,x,y,m}^I\cdot P_{S_{j,k,x,y}|C_{j,k,x,y},V_{j,k,i,l}}^I\left(m\middle| C_{j,k,x,y},V_{j,k,x,y}=v\right),
\end{split}
\label{eq:23}
\end{equation}
\end{small}
where ${Var}_{j,k,x,y,m}^I$ is the variance of $C_{j,k,x,y}$ in the hidden state $m$.

$P_{S_{j,k,x,y}|C_{j,k,x,y},V_{j,k,x,y}}^I(m|C_{j,k,x,y},V_{j,k,x,y}=v)$ represents the probability of the hidden state $m$ given the current coefficient and soft context variable $v$.

Then, the detail variable ${VPM}_{j,k,x,y,m}^I$ is defined based on ${VP}_{j,k,x,y,m}^I$:
\begin{equation}
\begin{split}
{VPM}_{j,k,x,y,m}^A=\left\{\begin{matrix}1,\ \ {VP}_{j,k,x,y,m}^A-{VP}_{j,k,x,y,m}^B\geq a v g\left(\left|{VP}_{j,k,m}^A-{VP}_{j,k,m}^B\right|\right)\\0,\ \ {VP}_{j,k,x,y,m}^B-{VP}_{j,k,x,y,m}^A\geq a v g\left(\left|{VP}_{j,k,m}^A-{VP}_{j,k,m}^B\right|\right)\\0.5+0.5\ast\frac{{VP}_{j,k,x,y,m}^A-{VP}_{j,k,x,y,m}^B}{mean\left(\left|{VP}_{j,k,m}^A-{VP}_{j,k,m}^B\right|\right)},\ \ otherwise\\\end{matrix}\right.
\end{split}
\label{eq:24}
\end{equation}

\begin{equation}
\begin{split}
{VPM}_{j,k,x,y,m}^B=1-{VPM}_{j,k,x,y,m}^A, 
\end{split}
\label{eq:25}
\end{equation}

where $avg(\cdot)$ represents the average value and $\vert \cdot \vert$ represents the absolute value.

If ${VP}_{j,k,x,y,m}^A>{VP}_{j,k,x,y,m}^B$ and their difference is great than or equal to the threshold,
${VPM}_{j,k,x,y,m}^A$ is recorded as 1.

On the contrary, if ${VP}_{j,k,x,y,m}^A<{VP}_{j,k,x,y,m}^B$ and their difference is great than or equal to the threshold, ${VPM}_{j,k,x,y,m}^A$ is recorded as 0. When the difference is less than the threshold, ${VPM}_{j,k,x,y,m}^A$ is between 0 and 1.

Afterwards, the multi-state saliency ${SM}_{j,k,x,y}^I$ is calculated by Eq.(\ref{eq:26}). The larger the ${SM}_{j,k,x,y}^I$ is, the higher the saliency of coefficient in all states is, otherwise, the smoother the coefficient is:

\begin{equation}
\begin{split}
{SM}_{j,k,x,y}^I=\frac{\sum_{m=0}^{n-1}{VPM}_{j,k,x,y,m}^I\cdot d{\left(m-\frac{n-1}{2}\right)}}{\sum_{m=0}^{n-1}d{\left(m-\frac{n-1}{2}\right)}},
\end{split}
\label{eq:26}
\end{equation}

\begin{equation}
\begin{split}
d{\left(u\right)}=\frac{1}{1+e^{-2u}}.
\end{split}
\label{eq:27}
\end{equation}

\begin{figure}[htb]
	\def\tempwidth{0.5\textwidth}
	\tiny
	\centering
	\subfigure{\includegraphics[width=\tempwidth]{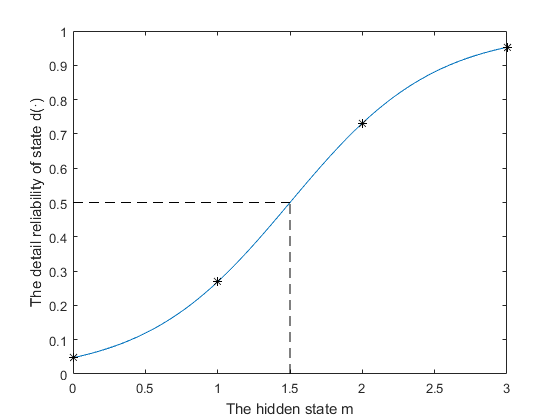}}
	\centering
	\caption{The graph of $d(\cdot)$ function with $n=4(m=0, 1, 2, 3)$. The discrete hidden state values are marked with * in this figure.}
	\label{fig:d_fuc}
\end{figure}

In the proposed NSST-MCHMM, the total number of states n divides the detail of the coefficient into n levels, which correspond to different hidden states, and the different hidden state can contribute differently to the information amount contained in the coefficients. Thus, the contribution of hidden state m for the current coefficients should be regarded as being in the corresponding detail interval respectively. The small hidden state captures few details, and the coefficients with large hidden states have rich details. Then, the $d(\cdot)$ function is proposed and defined as the detail reliability function of hidden state. Fig.\ref{fig:d_fuc} is the function graph of $d{\left(\cdot\right)}$ with the total number of states n=4 as an example. The abscissa is the hidden state, and the ordinate is the detail reliability under different states. The $d(\cdot)$ function can adaptively adjust the contributions of ${VPM}_{j,k,x,y,m}^I$ to get a more accurate ${SM}_{j,k,x,y}^I$. 

\begin{figure}[htb]
	\def\tempwidth{0.8\textwidth}
	\tiny
	\centering
	\subfigure{\includegraphics[width=\tempwidth]{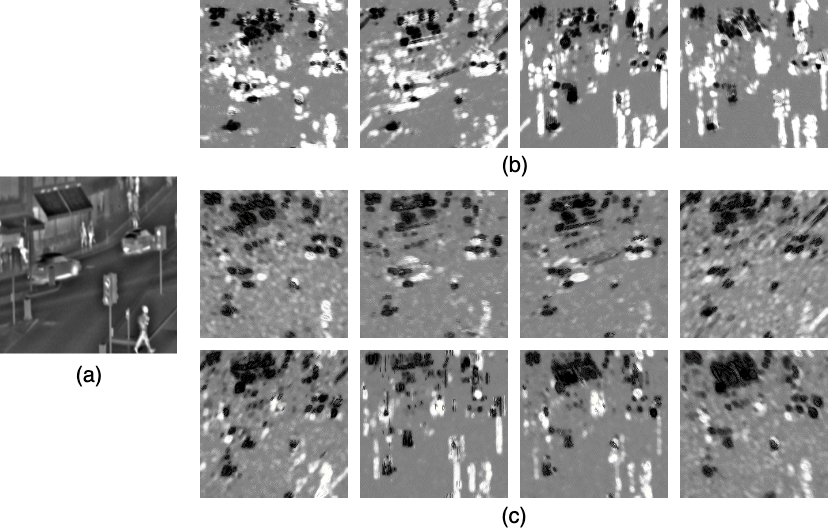}}
	\centering
	\caption{The visualized maps of ${SM}_{j,k}^I$. (a) the visible image; (b) the visualized maps of ${SM}_{j,k}^I(j=1;k=1,2,3,4)$; (c) the visualized maps of  ${SM}_{j,k}^I(j=2;k=1,2,\ldots\ldots,8)$.}
	\label{fig:VoSM}
\end{figure}

The visualization of multi-state saliency ${SM}_{j,k}^I$ is shown in Fig. \ref{fig:VoSM} by projecting the value into the gray space. The coefficient with low multi-state saliency displays black in the visualized map, i.e. the billboard of the store, which means that the coefficient here is lack of details. When the multi-state saliency of the coefficient is large, the pixel corresponding to this coefficient appears white in the visualized map, i.e. the traffic lights and pedestrians, which matches the details in the source image Fig. \ref{fig:VoSM}(a). The above analysis shows that the multi-state saliency ${SM}_{j,k}^I$ has the capability to extract the salient information of the source image.

The activity-level measure ${MH}_{j,k,x,y}^I$ is defined by combining the ${SM}_{j,k,x,y}^I$ with the $V_{j,k,x,y}^I$:

\begin{equation}
\begin{split}
{MH}_{j,k,x,y}^I=\alpha\ast{SM}_{j,k,x,y}^I+(1-\alpha)\ast V_{j,k,x,y}^I,
\end{split}
\label{eq:28}
\end{equation}
where $\alpha$ is used to balance the importance of ${SM}_{j,k,x,y}^I$ and $V_{j,k,x,y}^I$ and is set to 0.5.

Finally, the fused high-frequency subbands are obtained by the choose-max fusion strategy.

\begin{equation}
\begin{split}
C_{j,k,x,y}^F=\left\{\begin{matrix}C_{j,k,x,y}^A,{MH}_{j,k,x,y}^A\geq{MH}_{j,k,x,y}^B\\C_{j,k,x,y}^B,{MH}_{j,k,x,y}^A<{MH}_{j,k,x,y}^B\\\end{matrix}\right.
\end{split}
\label{eq:28}
\end{equation}

\section{Experimental results and discussions} \label{sec:exp}
To verify the effectiveness of proposed method, 21 sets of infrared and visible images \cite{19Ma2017Infrared} are adopted for fusion experiments. The images are all resized to the same size (256 * 256) to facilitate multi-scale transformation. All experiments are conducted in MATLAB R2015b on PC with 3.30GHz Inter(R) Core(TM) i5 CPU with 8 GB RAM. For comparison, several recent and classical fusion methods are selected including the cross bilateral filter based fusion method(CBF) \cite{20kumar2015Image}, the deep convolutional neural network based method (CNN) \cite{21Liu2017Multi}, the dense block based fusion method (DenseFuse) \cite{22Hui2018DenseFuse}, the GAN based fusion method (FusionGAN) \cite{23Ma2019FusionGAN}, the gradient transfer based fusion method(GTF) \cite{24Ma2016Infrared}, the joint-sparse representation model based fusion method (JSR) \cite{25Zhang2013Dictionary}, the JSR model with saliency detection based fusion method(JSRSD) \cite{26Liu2017Infrared}, the VGG-19 and multi-layer fusion strategy based method (VggML)  \cite{27li2018infrared} and the ResNet50 and zero-phase component analysis based fusion method (ZCA) \cite{28LI2019103039}. In the DenseFuse method, an l1-Norm strategy is applied to the network. The fusion framework uses the l1-Norm strategy and the convolutional block is Conv4 in the ZCA method. In the proposed algorithm, the contextual correlation weight is set as $\omega_0=0.8$, $\omega_1=0.6$, $\omega_2=0.2$,  $\omega_3=0.4$, the NSST decomposes two scales with ‘maxflat’ filter, and the number of high-frequency subband directions is 4, 8 respectively.

Ten objective evaluation indices are used to quantitatively compare the proposed fusion method with the other existing fusion methods. Qabf \cite{29Xydeas2000Objective} reflects the total information transferred from the source images to the fused image. FMI\_w \cite{30M2014Fast} is a fast mutual information index of wavelet feature. SCD \cite{31V2015A} calculates the correlation between the differential images computed on the source images and the fused image. MS-SSIM \cite{32ma2015perceptual} reflects the multi-scale structural similarity between the source images and the fused image. Average gradient (AG) is used to measure the clarity of the fused image. Edge intensity (EI) reflects the sharpness of the edge of the fused image. Q, Qw, and Qe are Piella metrics \cite{33G2003A}, which utilize local measures to estimate the salience of the fused image. Qw gives more weight to the window with higher saliency, while Qe pays more attention to the edge information. Spatial frequency (SF) reflects the change rate of gray level of image. For all the indices, a higher value corresponds to better performance.

Due to the limited space, three sets of infrared and visible image pairs are selected to illustrate the effectiveness of the proposed method. The average values of evaluation indices for 21 groups of images are given in the end.

\subsection{The first group of infrared and visible image fusion experiment}

\begin{figure}[htb]
	\addtolength{\abovecaptionskip}{-5pt}
	\addtolength{\belowcaptionskip}{-10pt}
	\def\tempwidth{0.16\textwidth}
	\tiny
	\centering
	\subfigure[]{\includegraphics[width=\tempwidth]{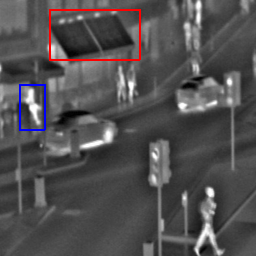}}
	\,
	\subfigure[]{\includegraphics[width=\tempwidth]{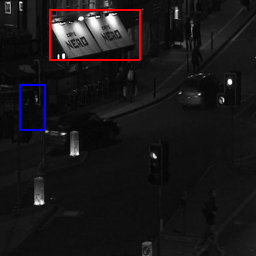}}
	\,
	\subfigure[]{\includegraphics[width=\tempwidth]{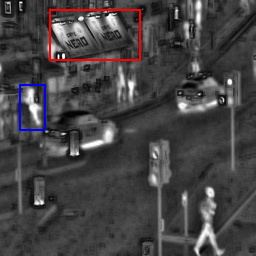}}
	\,
	\subfigure[]{\includegraphics[width=\tempwidth]{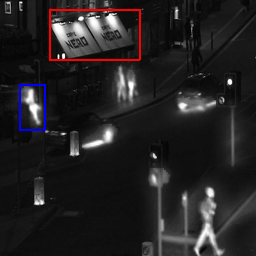}}
    \\
	\subfigure[]{\includegraphics[width=\tempwidth]{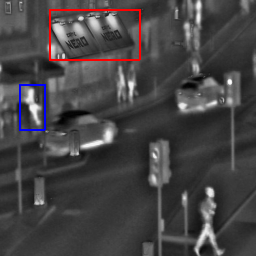}}
	\,
	\vspace{-2mm}
	\subfigure[]{\includegraphics[width=\tempwidth]{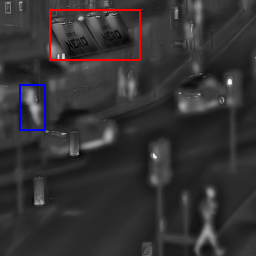}}	
	\,
	\subfigure[]{\includegraphics[width=\tempwidth]{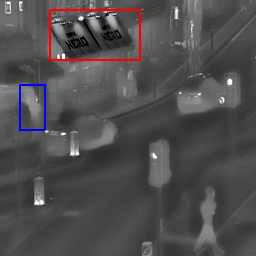}}
	\,
	\subfigure[]{\includegraphics[width=\tempwidth]{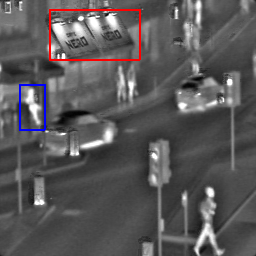}}
	\\
	\subfigure[]{\includegraphics[width=\tempwidth]{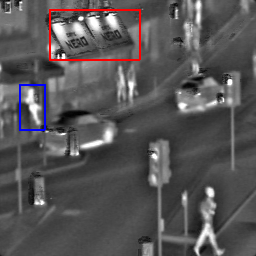}}
	\,
	\subfigure[]{\includegraphics[width=\tempwidth]{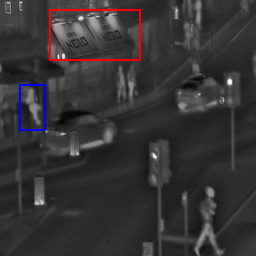}}
		\,
	\subfigure[]{\includegraphics[width=\tempwidth]{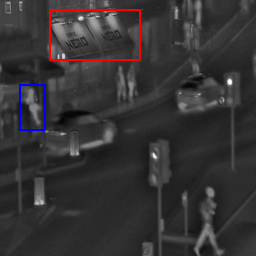}}
		\,
	\subfigure[]{\includegraphics[width=\tempwidth]{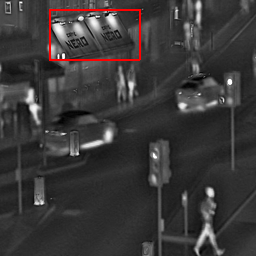}}
	\\
	\subfigure[]{\includegraphics[width=\tempwidth]{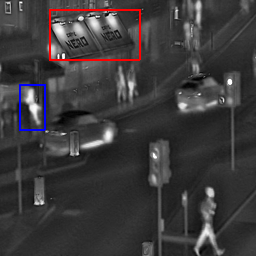}}
	\,
	\subfigure[]{\includegraphics[width=\tempwidth]{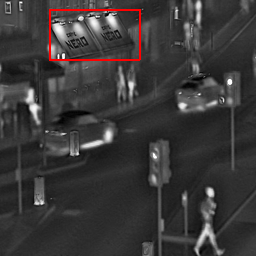}}
		\,
	\subfigure[]{\includegraphics[width=\tempwidth]{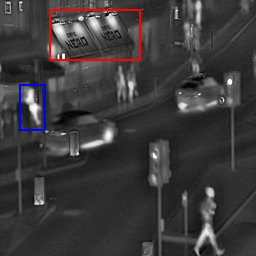}}
		\,
	\subfigure[]{\includegraphics[width=\tempwidth]{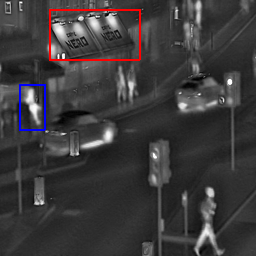}}
	
	\centering
	\caption{ The fusion experiment on “traffic” images. (a) Infrared image; (b) Visible image; (c) CBF; (d) CNN; (e) DenseFuse; (f) FusionGAN; (g) GTF; (h) JSR; (i) JSRSD; (j) VggML; (k) ZCA; (l)-(p) the proposed method with different number of states, which are $n=2, n=3, n=4, n=5, n=6$.}
	\label{fig:exp1}
\end{figure}

Fig. \ref{fig:exp1} shows the fusion experiment results of “traffic” infrared and visible images. Fig. \ref{fig:exp1}(a) and (b) are an infrared image and a visible image respectively. Fig. \ref{fig:exp1}(c)-(p) are the fused images obtained by different fusion methods. From a subjective point of view, the fused image based on CBF contains a lot of noise, which affects the clarity of the image. The CNN-based fused image is overall dark, so that the roadside shops and utility poles are difficult to identify. The fused images obtained by the GTF and FusionGAN lose the saliency and gradient information from infrared image (i.e., blue boxes in Fig.\ref{fig:exp1} (f) and Fig.\ref{fig:exp1} (g)). It is difficult to identify pedestrians at the door of the store. The fused images obtained by JSR and JSRSD suffer from unpleasant shadows near the edges. The fused images obtained by VggML and ZCA have low contrast and lack the prominent target information. The DenseFuse and proposed methods show better visual results compared to other methods. However, the DenseFuse method is weak in detail performance, i.e., the letters in the red box area cannot be clearly recognized. The proposed method performs well in detail and significant target information. In summary, compared with the comparative experiments, the proposed method has rich details, clear edge textures, and more prominent target information.

\begin{table}[htbp]
  \centering
  \caption{Objective evaluation for traffic images. }
   \resizebox{\textwidth}{!}{ \begin{tabular}{p{5em}c|cccccccccc}
    \hline
    \multicolumn{2}{p{10em}|}{Methods} & \multicolumn{1}{p{5em}}{Qabf} & \multicolumn{1}{p{5em}}{FMI\_w} & \multicolumn{1}{p{5em}}{SCD} & \multicolumn{1}{p{5em}}{MS\_SSIM} & \multicolumn{1}{p{5em}}{AG} & \multicolumn{1}{p{5em}}{EI} & \multicolumn{1}{p{5em}}{Q} & \multicolumn{1}{p{5em}}{Qw} & \multicolumn{1}{p{5em}}{Qe} & \multicolumn{1}{p{5em}}{SF} \\
    \hline
    \multicolumn{2}{p{10em}|}{DenseFuse} & \textbf{0.6618} & 0.4045 & 1.6984 & \textbf{0.9559} & 5.3919 & \textbf{51.7832} & \textbf{0.8466} & 0.8408 & 0.6142 & 13.1127 \\
    \multicolumn{2}{p{10em}|}{FusionGAN} & 0.2572 & 0.2998 & 1.2428 & 0.8344 & 3.2888 & 30.4994 & 0.6794 & 0.6099 & 0.2726 & 9.0979 \\
    \multicolumn{2}{p{10em}|}{GTF} & 0.3472 & 0.3971 & 1.0914 & 0.7512 & 3.776 & 32.3471 & 0.6613 & 0.52  & 0.2731 & 12.3178 \\
    \multicolumn{2}{p{10em}|}{VggML} & 0.4427 & 0.4153 & 1.5029 & 0.8666 & 3.7797 & 34.3889 & 0.7541 & 0.7088 & 0.4613 & 10.0285 \\
    \multicolumn{2}{p{10em}|}{ZCA} & 0.3964 & 0.4161 & 1.4597 & 0.8585 & 3.6065 & 32.9247 & 0.7483 & 0.6838 & 0.41  & 9.2594 \\
    \hline
    \multicolumn{1}{c|}{\multirow{5}[2]{*}{proposed}} & \multicolumn{1}{p{5em}|}{n=2} & \textit{0.6022} & 0.421 & 1.7617 & \textit{0.944} & 5.8467 & \textit{50.8725} & \textit{0.8112} & \textit{\textbf{0.8441}} & \textit{\textbf{0.6936}} & \textit{\textbf{15.5083}} \\
    \multicolumn{1}{c|}{} & \multicolumn{1}{p{5em}|}{n=3} & 0.6013 & \textit{\textbf{0.4224}} & 1.7615 & 0.9437 & 5.8576 & 50.8268 & 0.8111 & 0.8411 & 0.6832 & 15.3715 \\
    \multicolumn{1}{c|}{} & \multicolumn{1}{p{5em}|}{n=4} & 0.5997 & 0.4214 & 1.7616 & 0.9437 & 5.8552 & 50.7922 & 0.8111 & 0.841 & 0.6832 & 15.3797 \\
    \multicolumn{1}{c|}{} & \multicolumn{1}{p{5em}|}{n=5} & 0.5984 & 0.4215 & 1.7622 & 0.9436 & \textit{\textbf{5.8594}} & 50.833 & 0.811 & 0.8414 & 0.6849 & 15.4177 \\
    \multicolumn{1}{c|}{} & \multicolumn{1}{p{5em}|}{n=6} & 0.5974 & 0.4205 & \textit{\textbf{1.7623}} & 0.9436 & 5.8591 & 50.8167 & 0.8108 & 0.8414 & 0.6848 & 15.4204 \\
    \hline
    \multicolumn{2}{p{10em}|}{rank} & 2     & 1     & 1     & 2     & 1     & 2     & 2     & 1     & 1     & 1 \\
    \hline
    \end{tabular}%
    }
  \label{tab:T1}%
\end{table}%

When the visual perceptions of fusion results are close, objective evaluation is further needed. Considering the fact that the visual perception of Fig.\ref{fig:exp1}(c), Fig.\ref{fig:exp1}(d), Fig.\ref{fig:exp1}(i) and Fig.\ref{fig:exp1}(j) is obvious inferior, the following objective evaluation is executed on DenseFuse, FusionGAN, GTF, vggML, and ZCA methods and the proposed method. The objective evaluations for traffic fused images are given in Table 1, in which the last row is the rank of the proposed fusion method and six comparative experiments, the bold font is the best for all fusion results including comparative experiments, and the italic font is the best for the proposed fusion results with different numbers of states. As can be seen from Table \ref{tab:T1}, our method gets the best values in terms of FMI\_w, SCD, AG, Qw and Qe. It denotes that our method can retain the feature information of the source image while obtaining a fused image with high contrast. The proposed method is inferior to the DenseFuse method in Qabf, MS\_SSIM, EI, Q indices. However, the visual perception of DenseFuse is relatively blurred. In the case of different number of states, the proposed method with $n = 2$ has obvious advantages.

\subsection{The second group of infrared and visible image fusion experiment}

\begin{figure}[htb]
	\addtolength{\abovecaptionskip}{-5pt}
	\addtolength{\belowcaptionskip}{-10pt}
	\def\tempwidth{0.16\textwidth}
	\tiny
	\centering
	\subfigure[]{\includegraphics[width=\tempwidth]{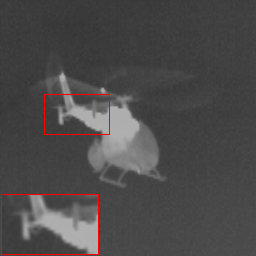}}
	\,
	\subfigure[]{\includegraphics[width=\tempwidth]{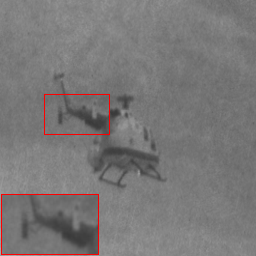}}
	\,
	\subfigure[]{\includegraphics[width=\tempwidth]{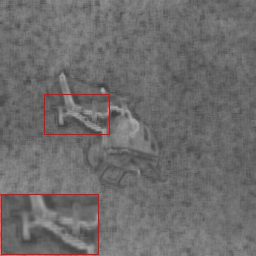}}
	\,
	\subfigure[]{\includegraphics[width=\tempwidth]{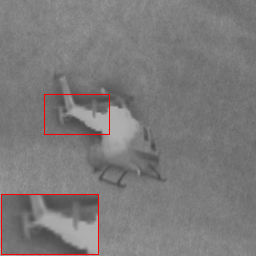}}
    \\
	\subfigure[]{\includegraphics[width=\tempwidth]{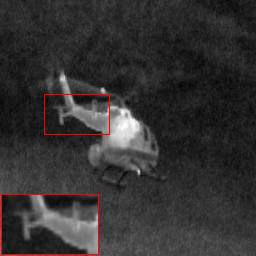}}
	\,
	\vspace{-2mm}
	\subfigure[]{\includegraphics[width=\tempwidth]{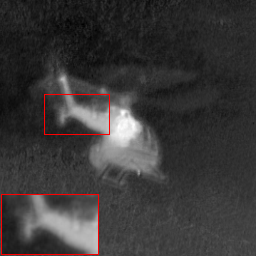}}	
	\,
	\subfigure[]{\includegraphics[width=\tempwidth]{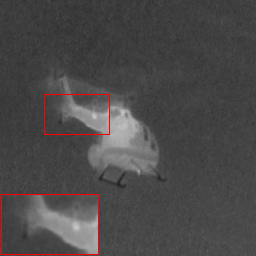}}
	\,
	\subfigure[]{\includegraphics[width=\tempwidth]{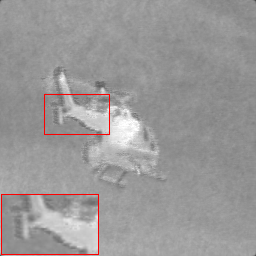}}
	\\
	\subfigure[]{\includegraphics[width=\tempwidth]{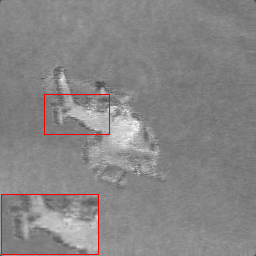}}
	\,
	\subfigure[]{\includegraphics[width=\tempwidth]{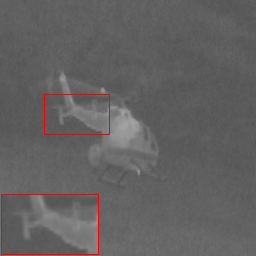}}
		\,
	\subfigure[]{\includegraphics[width=\tempwidth]{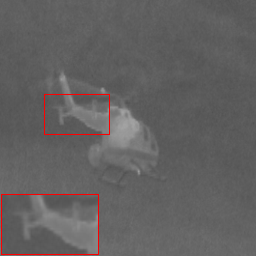}}
		\,
	\subfigure[]{\includegraphics[width=\tempwidth]{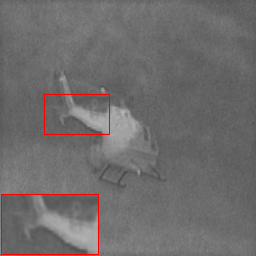}}
	\\
	\subfigure[]{\includegraphics[width=\tempwidth]{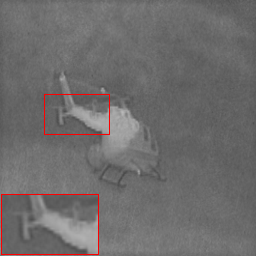}}
	\,
	\subfigure[]{\includegraphics[width=\tempwidth]{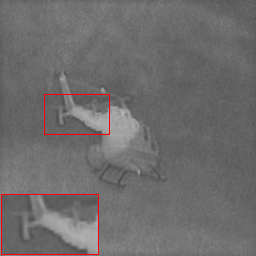}}
		\,
	\subfigure[]{\includegraphics[width=\tempwidth]{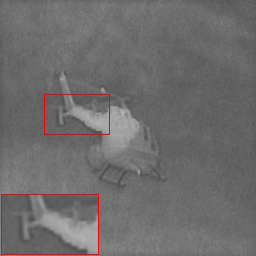}}
		\,
	\subfigure[]{\includegraphics[width=\tempwidth]{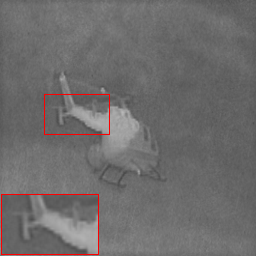}}
	
	\centering
	\caption{ The fusion experiment on “helicopter” images.  (a) Infrared image; (b) Visible image; (c) CBF; (d) CNN; (e) DenseFuse; (f) FusionGAN; (g) GTF; (h) JSR; (i) JSRSD; (j) VggML; (k) ZCA; (l)-(p) the proposed method with different number of states, which are $n=2, n=3, n=4, n=5, n=6$.}
	\label{fig:exp2}
\end{figure}

Fig. \ref{fig:exp2}(a) and (b) are the “helicopter” infrared and visible image, (c)-(p) are the experimental results obtained by different fusion methods, and the lower left corner of the image is the larger version of the red box area. In subjective perception, the fused images based on CBF, JSR and JSRSD all exhibit significant artificial noises, which affects the clarity of the image. The fused images based on GTF and FusionGAN are relatively blurred, i.e., the edges of the tail of the aircraft in the red box of Fig.\ref{fig:exp2}(f) and Fig.\ref{fig:exp2}(g). The fusion result obtained by the DenseFuse method has clear target information, but fails to maintain uniform gray in the background. The proposed method provides a more pleasing results compared with VggML, ZCA and CNN methods for the prominent target information and clear contours. To sum up, our method can get superior subjective perception. As can be seen from Fig.\ref{fig:exp2} (l)-(p), the edge of Fig.\ref{fig:exp2} (l) ($n = 2$) is blurred in the red box. The quality of Fig.\ref{fig:exp2} (m) is slightly better than Fig.\ref{fig:exp2} (l). When the number of states n of the proposed method is great than 3, the fused images maintain a clear edge. It can be seen that the proposed algorithm with multi-state has obvious advantages in preserving the detail of the source images.

\begin{table}[htbp]
  \centering
  \caption{Objective evaluation for helicopter images.}
   \resizebox{\textwidth}{!}{ \begin{tabular}{p{5em}c|cccccccccc}
    \hline
    \multicolumn{2}{p{10em}|}{Methods} & \multicolumn{1}{p{5em}}{Qabf} & \multicolumn{1}{p{5em}}{FMI\_w} & \multicolumn{1}{p{5em}}{SCD} & \multicolumn{1}{p{5em}}{MS\_SSIM} & \multicolumn{1}{p{5em}}{AG} & \multicolumn{1}{p{5em}}{EI} & \multicolumn{1}{p{5em}}{Q} & \multicolumn{1}{p{5em}}{Qw} & \multicolumn{1}{p{5em}}{Qe} & \multicolumn{1}{p{5em}}{SF} \\
    \hline
    \multicolumn{2}{p{10em}|}{DenseFuse} & 0.3763 & 0.3991 & 1.7015 & 0.8635 & \textbf{5.764} & \textbf{42.7932} & 0.6903 & 0.6631 & 0.38  & \textbf{10.2154} \\
    \multicolumn{2}{p{10em}|}{FusionGAN} & 0.3456 & 0.3941 & 1.1417 & 0.8845 & 3.6657 & 27.5394 & 0.623 & 0.642 & 0.2775 & 6.3716 \\
    \multicolumn{2}{p{10em}|}{GTF} & 0.5129 & \textbf{0.4635} & 1.3808 & 0.9245 & 2.6764 & 18.7185 & 0.8062 & 0.6448 & 0.3745 & 4.7264 \\
    \multicolumn{2}{p{10em}|}{VggML} & 0.4313 & 0.3927 & 1.6878 & 0.9181 & 1.7136 & 12.3768 & 0.892 & 0.6987 & 0.3517 & 3.0479 \\
    \multicolumn{2}{p{10em}|}{ZCA} & 0.4151 & 0.3921 & 1.6898 & 0.9161 & 1.657 & 12.0033 & 0.8898 & 0.6949 & 0.3415 & 2.9621 \\
    \hline
    \multicolumn{1}{c|}{\multirow{5}[2]{*}{proposed}} & \multicolumn{1}{p{5em}|}{n=2} & 0.5583 & \textit{0.4445} & 1.7559 & 0.9234 & 2.7422 & 19.2711 & 0.8963 & 0.6979 & 0.3868 & 4.7435 \\
    \multicolumn{1}{c|}{} & \multicolumn{1}{p{5em}|}{n=3} & 0.5635 & 0.442 & \textit{\textbf{1.7567}} & 0.925 & 2.7588 & 19.4313 & 0.898 & 0.7184 & 0.44  & 4.819 \\
    \multicolumn{1}{c|}{} & \multicolumn{1}{p{5em}|}{n=4} & \textit{\textbf{0.5692}} & 0.4412 & 1.7557 & \textit{\textbf{0.9257}} & \textit{2.7669} & \textit{19.5339} & \textit{\textbf{0.8989}} & \textit{\textbf{0.732}} & \textit{\textbf{0.487}} & \textit{4.8685} \\
    \multicolumn{1}{c|}{} & \multicolumn{1}{p{5em}|}{n=5} & 0.5675 & 0.4401 & 1.7562 & 0.9256 & 2.76  & 19.4881 & \textit{\textbf{0.8989}} & 0.7295 & 0.4756 & 4.8483 \\
    \multicolumn{1}{c|}{} & \multicolumn{1}{p{5em}|}{n=6} & 0.5668 & 0.44  & 1.7563 & 0.9255 & 2.762 & 19.4888 & 0.8987 & 0.7283 & 0.4707 & 4.8484 \\
    \hline
    \multicolumn{2}{p{10em}|}{rank} & 1     & 2     & 1     & 1     & 3     & 3     & 1     & 1     & 1     & 2 \\
    \hline
    \end{tabular}%
    }
  \label{tab:T2}%
\end{table}%

From the objective metrics in Table \ref{tab:T2}, the proposed method gets the maximum value in terms of Qabf, SCD, MS\_SSIM, Q, Qw, Qe. This fact illustrates that the proposed method has better capacity to transfer the detail into the fused image. The DenseFuse based method obtains excellent values on the AG, EI, and SF indices. The GTF-based method achieves the optimal value on the FMI index, but it performs poorly on the edge details. Within different state numbers, the proposed method with $n = 4$ is the best in terms of objective indicators, and also has good visual effects and contains useful information of the source images subjectively.

\subsection{The third group of infrared and visible image fusion experiment}

\begin{figure}[htb]
	\addtolength{\abovecaptionskip}{-5pt}
	\addtolength{\belowcaptionskip}{-10pt}
	\def\tempwidth{0.16\textwidth}
	\tiny
	\centering
	\subfigure[]{\includegraphics[width=\tempwidth]{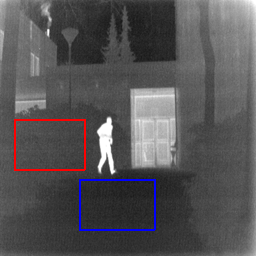}}
	\,
	\subfigure[]{\includegraphics[width=\tempwidth]{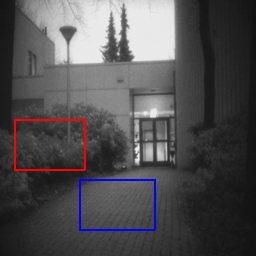}}
	\,
	\subfigure[]{\includegraphics[width=\tempwidth]{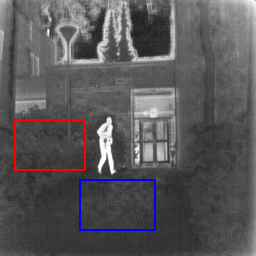}}
	\,
	\subfigure[]{\includegraphics[width=\tempwidth]{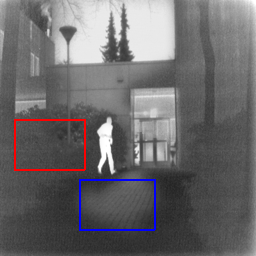}}
    \\
	\subfigure[]{\includegraphics[width=\tempwidth]{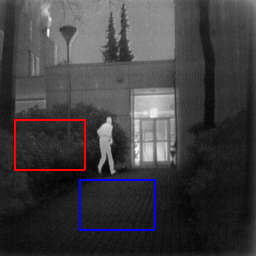}}
	\,
	\vspace{-2mm}
	\subfigure[]{\includegraphics[width=\tempwidth]{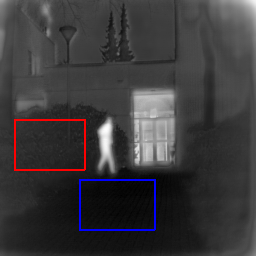}}	
	\,
	\subfigure[]{\includegraphics[width=\tempwidth]{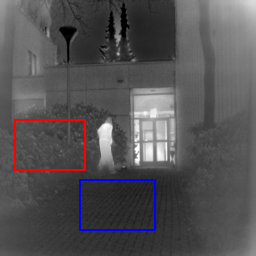}}
	\,
	\subfigure[]{\includegraphics[width=\tempwidth]{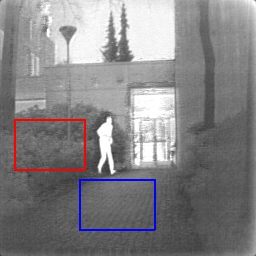}}
	\\
	\subfigure[]{\includegraphics[width=\tempwidth]{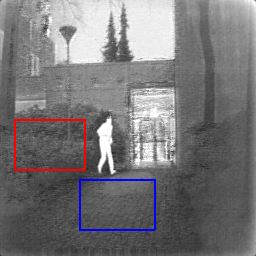}}
	\,
	\subfigure[]{\includegraphics[width=\tempwidth]{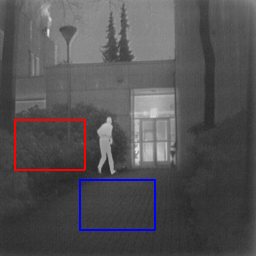}}
		\,
	\subfigure[]{\includegraphics[width=\tempwidth]{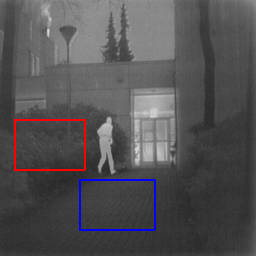}}
		\,
	\subfigure[]{\includegraphics[width=\tempwidth]{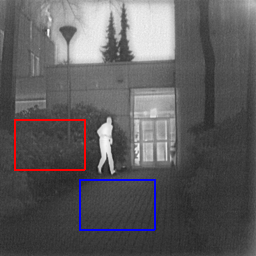}}
	\\
	\subfigure[]{\includegraphics[width=\tempwidth]{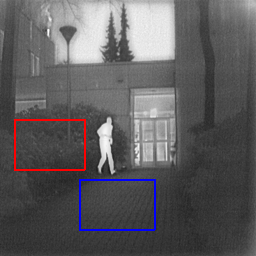}}
	\,
	\subfigure[]{\includegraphics[width=\tempwidth]{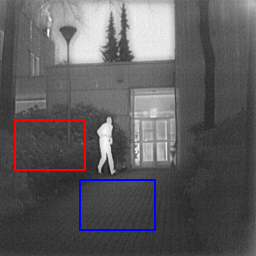}}
		\,
	\subfigure[]{\includegraphics[width=\tempwidth]{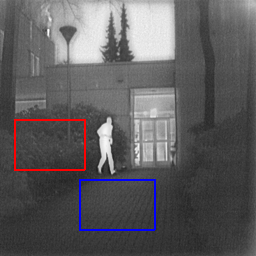}}
		\,
	\subfigure[]{\includegraphics[width=\tempwidth]{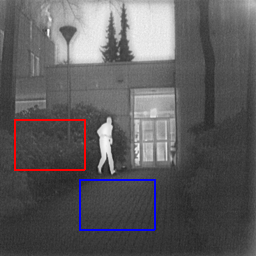}}
	
	\centering
	\caption{ The fusion experiment on “courtyard” images. (a) Infrared image; (b) Visible image; (c) CBF; (d) CNN; (e) DenseFuse; (f) FusionGAN; (g) GTF; (h) JSR; (i) JSRSD; (j) VggML; (k) ZCA; (l)-(p) the proposed method with different number of states, which are $n=2, n=3, n=4, n=5, n=6$.}
	\label{fig:exp3}
\end{figure}

Fig. \ref{fig:exp3} shows the fusion results of courtyard images, where Fig.\ref{fig:exp3} (a) and Fig. \ref{fig:exp3} (b) are infrared images and visible images, and Fig. \ref{fig:exp3} (c)-(p) are fused images obtained by different fusion methods. From a subjective point of view, the fused image obtained by CBF has poor visual quality and loses lots of visible image features. The fused images based on JSR and JSRSD have a lot of noise, especially in Fig. \ref{fig:exp3}(i), it degrades the edge information of the image, and the scene in the house cannot be seen clearly. The CNN-based and DenseFuse-based images fail to maintain uniform brightness, i.e., the road surface in Fig. \ref{fig:exp3} (d) and Fig. \ref{fig:exp3} (e). The fusion result based on FusionGAN is blurry overall, i.e., the edges of the silhouette are not clear. The GTF-based image lost outline information of the human from the infrared image. The proposed method produces a significantly enhanced fused image compared with the VggML and the ZCA. It is easy to see that the fused images obtained by the proposed method show the abundant texture of plant in the red box and clear ground pattern in the blue box. The qualities of fusion results obtained by the proposed method in different state numbers cannot be subjectively distinguished. The following objective quality evaluation is given to analyze them.

\begin{table}[htbp]
  \centering
  \caption{Objective evaluation for courtyard images.}
    \resizebox{\textwidth}{!}{\begin{tabular}{p{5em}c|cccccccccc}
    \hline
    \multicolumn{2}{p{10em}|}{Methods} & \multicolumn{1}{p{5em}}{Qabf} & \multicolumn{1}{p{5em}}{FMI\_w} & \multicolumn{1}{p{5em}}{SCD} & \multicolumn{1}{p{5em}}{MS\_SSIM} & \multicolumn{1}{p{5em}}{AG} & \multicolumn{1}{p{5em}}{EI} & \multicolumn{1}{p{5em}}{Q} & \multicolumn{1}{p{5em}}{Qw} & \multicolumn{1}{p{5em}}{Qe} & \multicolumn{1}{p{5em}}{SF} \\
    \hline
    \multicolumn{2}{p{10em}|}{DenseFuse} & \textbf{0.531} & 0.4252 & \textbf{1.7478} & \textbf{0.9078} & 5.0094 & 39.2809 & 0.7058 & 0.7924 & 0.6344 & 12.2536 \\
    \multicolumn{2}{p{10em}|}{FusionGAN} & 0.2349 & 0.3049 & 1.4979 & 0.6888 & 2.7355 & 23.0186 & 0.6302 & 0.5375 & 0.1983 & 6.7816 \\
    \multicolumn{2}{p{10em}|}{GTF} & 0.3818 & 0.407 & 0.803 & 0.7102 & 3.655 & 29.6755 & 0.7403 & 0.5496 & 0.304 & 9.3437 \\
    \multicolumn{2}{p{10em}|}{VggML} & 0.3799 & 0.424 & 1.5523 & 0.8575 & 3.4579 & 25.8769 & 0.8112 & 0.7079 & 0.4325 & 7.8705 \\
    \multicolumn{2}{p{10em}|}{ZCA} & 0.3599 & 0.4227 & 1.5537 & 0.8548 & 3.368 & 25.2278 & 0.808 & 0.6915 & 0.3988 & 7.5094 \\
    \hline
    \multicolumn{1}{r|}{\multirow{5}[2]{*}{proposed}} & \multicolumn{1}{p{5em}|}{n=2} & 0.5068 & 0.447 & 1.6854 & \textit{0.9056} & 5.499 & 39.5981 & \textit{\textbf{0.8131}} & 0.8219 & 0.6722 & 12.2689 \\
    \multicolumn{1}{r|}{} & \multicolumn{1}{p{5em}|}{n=3} & 0.5073 & 0.4519 & 1.6857 & 0.9055 & 5.5424 & 39.7322 & 0.8129 & 0.8222 & 0.6717 & 12.3469 \\
    \multicolumn{1}{r|}{} & \multicolumn{1}{p{5em}|}{n=4} & 0.5091 & 0.4525 & \textit{1.6859} & 0.9055 & 5.5566 & 39.8137 & 0.8129 & 0.8225 & 0.6725 & 12.3756 \\
    \multicolumn{1}{r|}{} & \multicolumn{1}{p{5em}|}{n=5} & \textit{0.5099} & 0.4531 & \textit{1.6859} & 0.9054 & 5.5682 & 39.8486 & 0.8127 & \textit{\textbf{0.8226}} & \textit{\textbf{0.6728}} & 12.3899 \\
    \multicolumn{1}{r|}{} & \multicolumn{1}{p{5em}|}{n=6} & 0.5096 & \textit{\textbf{0.4536}} & \textit{1.6859} & 0.9053 & \textit{\textbf{5.5749}} & \textit{\textbf{39.8656}} & 0.8125 & 0.8225 & 0.6726 & \textit{\textbf{12.3983}} \\
    \hline
    \multicolumn{2}{p{10em}|}{rank} & 2     & 1     & 2     & 2     & 1     & 1     & 1     & 1     & 1     & 1 \\
    \hline
    \end{tabular}%
    }
  \label{tab:T3}%
\end{table}%

The objective evaluation indices are listed in Table \ref{tab:T3}. The proposed method obtains the best performance in most evaluation indicators. This indicates that the feature information of inputs can be preserved by the proposed method better than other methods. Therefore, the fused image obtained by the proposed method has richer details and better visual effects. Under different number of states, the optimal objective evaluation values of the fusion results almost are obtained when n=6. In terms of subjective and objective evaluation, the quality of the proposed method is the best.

\subsection{21 sets of infrared and visible image fusion experiments}

\begin{table}[htbp]
  \centering
  \caption{The average values of objective evaluation for 21 fused images}
    \resizebox{\textwidth}{!}{ \begin{tabular}{p{5em}|cccccccccc}
    \hline
    Methods & \multicolumn{1}{p{5em}}{Qabf} & \multicolumn{1}{p{5em}}{FMI\_w} & \multicolumn{1}{p{5em}}{SCD} & \multicolumn{1}{p{5em}}{MS\_SSIM} & \multicolumn{1}{p{5em}}{AG} & \multicolumn{1}{p{5em}}{EI} & \multicolumn{1}{p{5em}}{Q} & \multicolumn{1}{p{5em}}{Qw} & \multicolumn{1}{p{5em}}{Qe} & \multicolumn{1}{p{5em}}{SF} \\
    \hline
    DenseFuse & 0.4911 & 0.4   & \textbf{1.7138} & 0.8327 & 5.5773 & 45.0983 & 0.7387 & 0.7213 & 0.486 & 11.9516 \\
    FusionGAN & 0.2545 & 0.3289 & 1.4554 & 0.6922 & 4.0368 & 31.5716 & 0.6196 & 0.5347 & 0.1926 & 8.1829 \\
    GTF   & 0.4337 & \textbf{0.4411} & 1.0388 & 0.744 & 5.2186 & 40.8233 & 0.7225 & 0.5628 & 0.2965 & 11.1838 \\
    VggML & 0.3896 & 0.4146 & 1.6404 & 0.8429 & 3.8165 & 30.4426 & 0.7909 & 0.7   & 0.3842 & 7.9498 \\
    ZCA & 0.3697 & 0.4135 & 1.6351 & 0.8389 & 3.7065 & 29.6296 & 0.7869 & 0.6872 & 0.3573 & 7.5876 \\
    \hline
    proposed & \textbf{0.5028} & 0.4282 & 1.6766 & \textbf{0.8414} & \textbf{6.1363} & \textbf{46.61} & \textbf{0.794} & \textbf{0.7589} & \textbf{0.541} & \textbf{12.914} \\
    \hline
    rank  & 1     & 2     & 2     & 2     & 1     & 1     & 1     & 1     & 1     & 1 \\
    \hline
    \end{tabular}%
    }
  \label{tab:T4}%
\end{table}%

The MCHMM statistical model is not limited to the “large” and “small” states. The coefficient is divided into n levels between smoothness and edges to illustrate the degree of detail. With the increasing of the number of state, the accuracy of MCHMM is enhanced. However, it is considered that the time consumption of algorithm also increases. Thus, the selection of state number is important for the performance of the proposed fusion method. The optimal value of n should balance the fusion performance and time consumption.  According to the analyses of the above three groups of fusion experiments, the parameter of state n is defined as 4 in the 21sets of infrared-visual fusion experiments. Considering the limited space, only the average values of ten metrics for 21 fused images obtained by existing methods and the proposed fusion method are shown in Table \ref{tab:T4}. The proposed method achieves the best values in terms of Qabf, AG, EI, Q, Qw, Qe, SF indices, and the SCD, FMI\_w and MS\_SSIM indices of the proposed method are close to the optimal values. The experimental results show that our method retains the edge and detail information of the source images and has good contrast, which is in line with the results of subjective evaluation.

\subsection{Experimental analyses}
The CBF method obtains the detail strength of the source images according to the eigenvalue of the unbiased estimate of covariance matrix. The detail strength is used to define the fusion weight of detail layers. However, the definition of fusion weight fails in some smooth areas because only considering a single feature without other correlations will lead to the inappropriate interpretation of image feature, thus leading to inferior fusion results. The CNN method uses synthesized images by adding Gaussian blur on all-in-focus image to train the network. This trained network is not suitable for the feature extraction of infrared and visible images as it failed to recognize the multi-modality characteristics, resulting in the unnatural combination of source images. Moreover, only the last layer results are utilized for fusion, and a lot of useful information existing in the middle layers are lost. The VggML method uses VGG-19 to extract multi-layer network features. The ZCA method uses zero-phase component analysis to normalize the depth features extracted from the residual network to get the initial weight maps, and adopts the weighted-averaging strategy to reconstruct the fused image. Both the above two methods need to resize the initial weight maps to the size of the source image through the up-sampling operations, which reduce the accuracy of weight for image fusion, and results in blurry fused images. In addition, the average fusion strategy for base layer will cause the degradation of image contrast. The JSR and JSRSD methods are based on joint sparse representation, and the JSRSD method improves the JSR method by taking the salient information of source images into consideration to get a better subjective effect. However, as JSR and JSRSD take a sliding-window technology to divide images into overlapping patches, and each patch is decomposed independently, their representation for image is multi-valued. Thereby, the aggregation of these patch to get the final value can inevitably distort the local structure in source images. The GTF method constrains the fused image to have similar pixel intensity distribution with the given infrared image and similar pixel gradients with the visible image. The FusionGAN method uses the generator to generate a fused image with major infrared intensities and additional visible gradients, and applies discriminator to force the fused image to have more details existing in visible image. Therefore, the fused images based on GTF and FusionGAN look like infrared images but lose the details existing in the visible image, and have an inferior visual perception due to the low resolution of infrared images, therefore the values of EI and SF indexes which measure the edge and gray change of fused images are low respectively. The DenseFuse method adds dense blocks to the encoder, and the dense connections ensures that the output of each layer can be directly used by subsequent layers, thus more useful features can be obtained for image fusion and the fused image obtains high values in Qabf, MS\_SSIM and SF indexes. However, the DenseFuse method uses MS-COCO dataset to train the feature extraction and reconstruction ability of the encoder and decoder, enforcing the trained network to extract the same features from infrared and visible image, weakening the details of fused image. The proposed method integrates both the target information from the infrared image and the detailed information from the visible image, and obtains a clear fused image with abundant information. The reason is that the low-frequency fusion of the proposed method is based on the difference of regional energy, which can enhance the contrast of the fused image and avoid introducing false information. The novel NSST-MCHMM model established on the high-frequency subbands is a fine-grained statistical model, and the training process from global to local improves the credibility of the model. The proposed soft context scheme accurately defines the details of the coefficients from the perspective of multiple correlations. Therefore, the statistical characteristics of the model not only include the global and local characteristics of the coefficients, but also reflect the contextual information of the coefficient, which is a comprehensive and accurate expression. On this basis, two novel statistical features, namely multi-state saliency and soft context variable of coefficient, are extracted as the activity level measurement for the fusion of high-frequency coefficients. Therefore, the fused image obtained by the proposed method can combine the complementary information from the infrared and visible images with high contrast, and acquires the highest values for most objective indices.

\section{Acknowledge:}
This work was supported in part by the National Natural Science Foundation of China under Grant 61772237, in part by the Six Talent Climax Foundation of Jiangsu under Grant XYDXX-030.

%
%
\bibliographystyle{unsrt} 
\bibliography{egbib}

\end{document}